\documentclass[letterpaper, 10 pt, conference]{ieeeconf}

\IEEEoverridecommandlockouts
\usepackage{amsmath,amsfonts,amssymb}
\usepackage{array}
\usepackage{booktabs}
\usepackage{cite}
\usepackage{adjustbox}
\usepackage{graphicx}
\usepackage{makecell}
\usepackage{multirow}
\usepackage{textcomp}
\usepackage{url}
\usepackage{xspace}
\usepackage[table,dvipsnames]{xcolor}
\usepackage[colorlinks=true,linkcolor=blue,citecolor=green,urlcolor=blue]{hyperref}

\definecolor{mildcolor}{RGB}{22,105,122}
\newcommand{\mild}{MILD\xspace}
\newcommand{\mildsim}{MILD-Sim\xspace}
\newcommand{\aprilvins}{AprilVINS\xspace}
\newcommand{\aprilvinsfull}{AprilVINS(full)\xspace}
\newcommand{\aprilvinsnotag}{AprilVINS W/O Tag Factors\xspace}
\newcommand{\aprilvinsnopromo}{AprilVINS W/O Output Gate\xspace}
\newcommand{\best}[1]{\cellcolor{Green!25}\textbf{#1}}
\newcommand{\secondbest}[1]{\cellcolor{SpringGreen!45}\textit{#1}}
\newcommand{\thirdbest}[1]{\cellcolor{Yellow!30}#1}

\makeatletter
\long\def\@makecaption#1#2{%
\ifx\@captype\@IEEEtablestring%
\setbox\@tempboxa\hbox{\footnotesize #1.~~\scshape #2}%
\ifdim \wd\@tempboxa >\hsize%
\setbox\@tempboxa\hbox{\footnotesize #1.~~ }%
\parbox[t]{\hsize}{\footnotesize\centering\noindent\unhbox\@tempboxa\scshape #2}%
\else%
\begin{center}{\footnotesize #1.~~\scshape #2}\end{center}%
\fi%
\@IEEEtablecaptionsepspace%
\else%
\@IEEEfigurecaptionsepspace%
\setbox\@tempboxa\hbox{\footnotesize #1.~~ #2}%
\ifdim \wd\@tempboxa >\hsize%
\setbox\@tempboxa\hbox{\footnotesize #1.~~ }%
\parbox[t]{\hsize}{\footnotesize \noindent\unhbox\@tempboxa#2}%
\else%
\ifcenterfigcaptions \hbox to\hsize{\footnotesize\hfil\box\@tempboxa\hfil}%
\else \hbox to\hsize{\footnotesize\box\@tempboxa\hfil}%
\fi\fi\fi}
\makeatother

\title{\LARGE \bf
Towards Accurate End-Effector Localization for UMI-Style Robotic Manipulation Teaching
}

\author{
Junjie Zhang$^{1,\dagger,\ddagger}$, Deteng Zhang$^{2,\ddagger}$, Zhisong Xu$^{3}$, Bo Sun$^{4}$, Liuyang Li$^{5}$, Yihong Tian$^{6}$, Jie Yin$^{7}$* \\
\thanks{$^{\dagger}$The work is partially done in PokeBot, $^*$ Corresponding author, $^{\ddagger}$Equal contribution}%
\thanks{$^{1}$Chongqing University, $^{2}$Northwestern Polytechnical University, $^{3}$The University of Tokyo, $^{4}$Jiaxing University, $^{5}$Sichuan University, $^{6}$Beijing Institute of Technology, $^{7}$Shanghai Jiao Tong University}%
}

\makeatletter
\IEEEaftertitletext{%
\vspace{-2.2\baselineskip}
\begin{center}
\textbf{Project Page: \url{https://mild-web.github.io/}}
\end{center}
\vspace{6pt}%
}
\makeatother

\begin{document}

\maketitle
\thispagestyle{empty}
\pagestyle{empty}

\begin{abstract}
Recovering end-effector trajectories for robot demonstration learning requires localization to remain accurate and temporally complete as a gripper approaches objects, occludes the camera, and interacts with the workspace. Existing SLAM benchmarks emphasize navigation motions, whereas manipulation datasets prioritize policy learning over localization evaluation. We introduce \mild, a Manipulation-Interface Localization Dataset with real-world and simulation sequences. The real-world subset provides 86 sensor sequences from Insta360 X5 and Insight9 across 15 repeated tabletop tasks, calibration assets, and a per-execution robot end-effector reference trajectory. The simulation subset, \mildsim, extends task coverage in Isaac Sim for controlled manipulation-replay studies. Benchmarking visual--inertial and fiducial-aided systems on instrumented real-world recordings reveals large differences in both TCP-relative trajectory error and temporal coverage, even under the same nominal task. To support marker-augmented teaching workspaces without a pre-surveyed fiducial map, we present \aprilvins, which combines fisheye visual--inertial estimation with sequence-local AprilTag geometry and separates prior admission from guarded export of the jointly optimized state. On Insta360 AprilTag4 recordings, \aprilvinsfull under a unified protocol with sequence-specific profiles reaches millimeter-level SE(3)-aligned TCP-relative APE RMSE with high time completion and lower reported error than the tested routes under their respective protocols, whereas fisheye VIO without tag factors remains at centimeter scale. Ablations separate accuracy from exportability, and a \mildsim replay study provides task-specific tolerance references for interpreting those error magnitudes. Together, \mild and \aprilvins provide a diagnostic benchmarking framework for UMI-style demonstration collection. Code, datasets, and evaluation manifests will be released upon acceptance.

\noindent\textbf{Keywords---} Manipulation-interface localization, visual--inertial odometry, AprilTag, robot teaching, benchmark dataset.
\end{abstract}

\section{Introduction}
\label{sec:introduction}

Demonstration collection for robot manipulation requires recovering the demonstrated end-effector trajectory from a portable manipulation interface. UMI systems have made in-the-wild data acquisition practical through a map-then-localize pipeline~\cite{chi2024umi}, but close-range manipulation stresses localization: hands and objects occlude the camera, useful features leave the field of view near contact, and slow motions can break visual--inertial initialization. These conditions resemble the degradation cases studied in ground-robot SLAM~\cite{yin2023ground,yin2024ground}, but at tabletop scale with tighter coverage requirements. A low-error fragment is not enough when the rest of a demonstration is untracked, so manipulation-interface localization should report both trajectory accuracy and temporal coverage.

Existing localization benchmarks target aerial, hand-held, and mobile-robot motion~\cite{burri2016euroc,schubert2018tum,yin2021m2dgr,zhang2025towards}, whereas manipulation datasets focus on policy learning~\cite{khazatsky2024droid_dataset,fang2023rh20t,zhaxizhuoma2025fastumi,jin2026umibench}. \mild complements the latter by benchmarking the sensing layer that determines whether a collected demonstration is replayable, rather than evaluating downstream policy quality directly.

This gap matters because localization error enters a demonstration twice: it changes the recovered path used for learning and, during replay, changes the spatial relation between the gripper and the object. The relevant tolerance is task dependent. Free-space transport may absorb centimeter-scale offsets, whereas insertion, stacking, and sustained contact can fail after only a few millimeters. A useful benchmark should therefore connect estimator-level metrics to task-level consequences instead of treating aggregate trajectory error as the final criterion.

We introduce \mild with real-world and simulation subsets. An AGIbot G2 humanoid executes 15 tasks under uninstrumented, textured, and fiducial-marker workspace variants while Insta360 X5 fisheye and Insight9 stereo/RGB-D sensors record the motion. Each execution has its own TCP reference, so error is measured against the motion actually executed. \mildsim adds Isaac Sim tasks for controlled replay studies (Sec.~\ref{sec:exp_replay}). Benchmarking on \mild motivates \aprilvins (Sec.~\ref{sec:aprilvins}), which estimates sequence-local AprilTag geometry without a pre-surveyed map, inserts admitted tag priors into a fisheye VIO window, and guards export of the optimized trajectory.

We summarize the core contributions of this work as follows:
\begin{itemize}
\item We introduce a manipulation-interface localization benchmark named \mild for demonstration collection, featuring repeated tabletop tasks, multi-sensor recordings, scene variants, calibration assets, per-execution end-effector reference trajectories, and a paired simulation subset \mildsim.
\item We develop \aprilvins, an AprilTag-assisted fisheye VIO system that estimates sequence-local tag geometry without an externally surveyed map, inserts admitted tag priors into a sliding window, and guards export of the joint state; ablations show that tag factors drive most accuracy gains while the output gate restores export continuity.
\item We establish a multi-metric evaluation protocol---APE RMSE, time completion, and continuity---for vision-only, fiducial-aided, visual--inertial, and commercial demonstration-tracking routes, and use \mildsim replay curves as task-specific tolerance references for the reported error magnitudes.
\end{itemize}

\section{Related Work}

\subsection{Visual--Inertial Localization}
Modern VIO and visual SLAM systems combine feature tracking, inertial preintegration, and sliding-window optimization~\cite{forster2016manifold,qin2018vins,geneva2020openvins,campos2021orb,von2022dm}. Handheld interfaces already carry cameras and IMUs, and fisheye models are common on manipulation-mounted cameras~\cite{kannala2006generic}. Close-range demonstrations nonetheless break navigation assumptions: short trajectories, fast rotation near the workspace, repeated occlusion, and the need to convert camera poses into replayable end-effector trajectories.

\subsection{Localization Benchmarks and Multi-Sensor SLAM}
EuRoC and TUM-VI standardize visual--inertial evaluation~\cite{burri2016euroc,schubert2018tum}, while ground-robot datasets such as M2DGR, Ground-Challenge, MARS-LVIG, and FusionPortable broaden sensor coverage and degradation cases~\cite{yin2021m2dgr,yin2023ground,li2024mars,jiao2022fusionportable}. Recent fusion systems target visual degradation, LiDAR degeneracy, and GNSS denial~\cite{yin2024ground,zhang2025towards,tian2026ultrafusion,he2025ligo}. \mild complements these lines by benchmarking close-range manipulation-interface localization with repeated workspace variants and end-effector references. Manipulation introduces a different failure regime: the workspace is compact, but occlusion is frequent and trajectory completeness directly determines whether a demonstration can be replayed.

\subsection{Manipulation and UMI-Style Datasets}
Large-scale manipulation datasets such as DROID, BridgeData V2, RH20T, and Open X-Embodiment target policy learning~\cite{khazatsky2024droid_dataset,walke2023bridgedata,fang2023rh20t,openx2023rtx}. UMI-style interfaces reduce collection overhead~\cite{chi2024umi,zhaxizhuoma2025fastumi,jin2026umibench,wang2026umi3d} but rarely evaluate localization directly. \mild is intended as a complementary diagnostic benchmark for that collection stack.

\subsection{Fiducial-Aided Localization}
Fiducial markers provide six-degree-of-freedom pose observations from images~\cite{olson2011apriltag}. TagSLAM and UcoSLAM combine fiducials with factor-graph or feature-based pipelines~\cite{pfrommer2019tagslam,munoz2020ucoslam}. Marker visibility alone does not guarantee a replayable trajectory, because fisheye projection, map consistency, synchronization, and temporal coverage all matter under close-range manipulation. This motivates estimating tag geometry online and guarding the exported trajectory rather than treating every detected tag pose as ground truth.

\section{The MILD Dataset}
\label{sec:mild_dataset}

\begin{figure*}[t]
\centering
\includegraphics[width=0.98\textwidth]{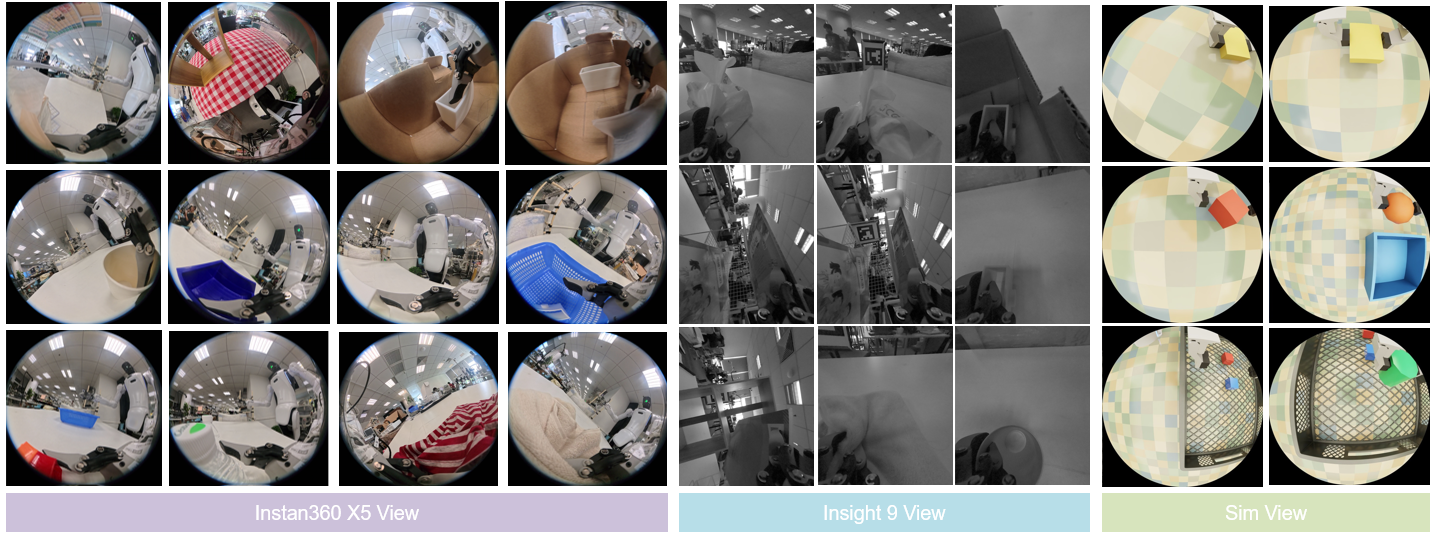}
\caption{Representative \mild recordings from Insta360 X5 (fisheye), Insight9 (stereo/RGB-D), and \mildsim (simulated fisheye) across pick-and-place, wiping, and rearrangement tasks. The montage shows the multi-sensor, cross-domain coverage of the benchmark.}
\label{fig:Scene_overview}
\end{figure*}

\subsection{Dataset Scope}
\mild targets end-effector trajectory recovery from a manipulation-mounted sensor stream. The evaluation unit is a task--scene--sensor recording with calibration and its own executed TCP reference. The reference comes from logged end-effector poses rather than external motion capture, so reported accuracy is relative to executed robot motion.

The benchmark follows three design choices. First, repeating a nominal task across scene and sensor variants separates task geometry from sensing conditions. Second, retaining raw observations and calibration permits evaluation with vision-only, visual--inertial, and fiducial-aided estimators. Third, every execution has its own reference rather than reusing the command trajectory: record-and-replay preserves the intended motion but does not guarantee identical realized motion.

\subsection{Real-World Data}
\begin{figure*}[t]
\centering
\includegraphics[width=0.98\textwidth]{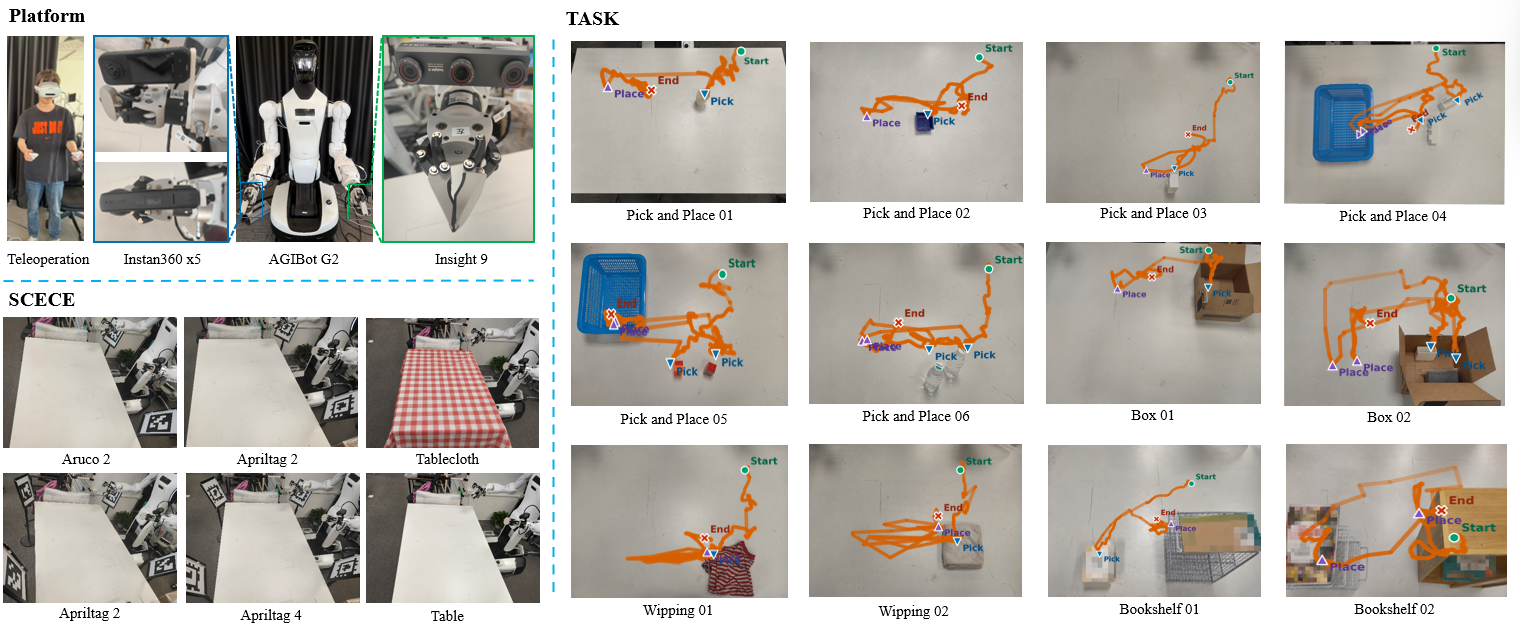}
\caption{\mild acquisition overview. \textbf{Platform:} AGIbot G2 with Insta360 X5 and Insight9 under teleoperation. \textbf{Scene:} plain tabletop, tablecloth, and ArUco/AprilTag marker layouts. \textbf{Task:} 15 tabletop families with executed TCP references; orange curves show trajectories and green/blue/red markers denote start, pick/place, and end events.}
\label{fig:mild_overview}
\end{figure*}

An AGIbot G2 humanoid executes 15 tabletop tasks: Analemma, Circular, Zigzag, Bookshelf 01--02, Box 01--02, Pick-and-place 01--06, and Wiping 01--02. The release contains 86 sequences (67 Insta360 X5, 19 Insight9) under uninstrumented, textured, and fiducial-marker workspace variants. Fig.~\ref{fig:Scene_overview} shows representative views from all three sensor domains, and Fig.~\ref{fig:mild_overview} summarizes the acquisition platform, scene variants, and annotated task trajectories. A record-and-replay procedure keeps nominal motion consistent: a reference demonstration logs joint commands, TCP poses, gripper feedback, and timestamps, and each replay yields a new sensor recording, calibration metadata, and executed TCP trajectory.

The variants serve distinct diagnostic roles. The uninstrumented tabletop tests natural-feature tracking under sparse close-range structure. The tablecloth changes texture without changing the nominal task, helping distinguish feature scarcity from motion-induced failure. ArUco and AprilTag layouts provide controlled observations with two or four markers. Their identities and sizes are known, but their global poses need not be surveyed. The marker recordings therefore also test sequence-local map construction.

Table~\ref{tab:mild_dataset_overview} summarizes the task inventory by family from the 86 indexed real-world recordings and 9 listed \mildsim entries.

\begin{table}[t]
\centering
\caption{MILD dataset overview by task family.}
\label{tab:mild_dataset_overview}
\scriptsize
\setlength{\tabcolsep}{2.0pt}
\renewcommand{\arraystretch}{1.12}
\resizebox{\columnwidth}{!}{%
\begin{tabular}{l r r r r r r r r r}
\toprule
\textbf{Item} & \textbf{Analemma} & \textbf{Circular} & \textbf{Zigzag} & \textbf{Bookshelf} & \textbf{Box} & \textbf{\makecell{Pick-and-\\place}} & \textbf{Wiping} & \textbf{Throwing}  & \textbf{Total} \\
\midrule
Real-world number & 9 & 9 & 11 & 15 & 7 & 23 & 12 & 0  & 86 \\
Simulation number & 1 & 1 & 1 & 0 & 0 & 4 & 1 & 1  & 9 \\
number & 10 & 10 & 12 & 15 & 7 & 27 & 13 & 1 & 95 \\
Time (min) & 5.4 & 4.9 & 17.9 & 31.0 & 16.0 & 41.0 & 28.9 & 0.2  & 145.3 \\
Size (GB) & 9.6 & 8.7 & 27.9 & 62.6 & 35.8 & 100.2 & 62.7 & 0.33  & 307.83 \\
\bottomrule
\end{tabular}
}
\begin{flushleft}
\footnotesize{Real-world counts are indexed recordings and simulation counts are listed \mildsim task entries. Time and size are accumulated from released real-world TUM and bag files; simulation totals will accompany the release.}
\end{flushleft}
\end{table}

Released calibration assets express camera or visual--inertial estimates in the robot TCP frame for both Insta360 X5 fisheye and Insight9 stereo/RGB-D routes. Each task--scene--sensor recording is evaluated against its own executed TCP reference because replays share nominal commands but not identical measured motion.

The reference is derived from logged robot kinematics and encoder feedback, so TCP-relative error reflects the trajectory available to the teaching pipeline together with residual calibration, synchronization, and robot-state uncertainty. We report SE(3)-aligned APE RMSE in this frame throughout the experiments.

\subsection{Simulation Data}
\begin{figure}[t]
\centering
\includegraphics[width=0.92\columnwidth]{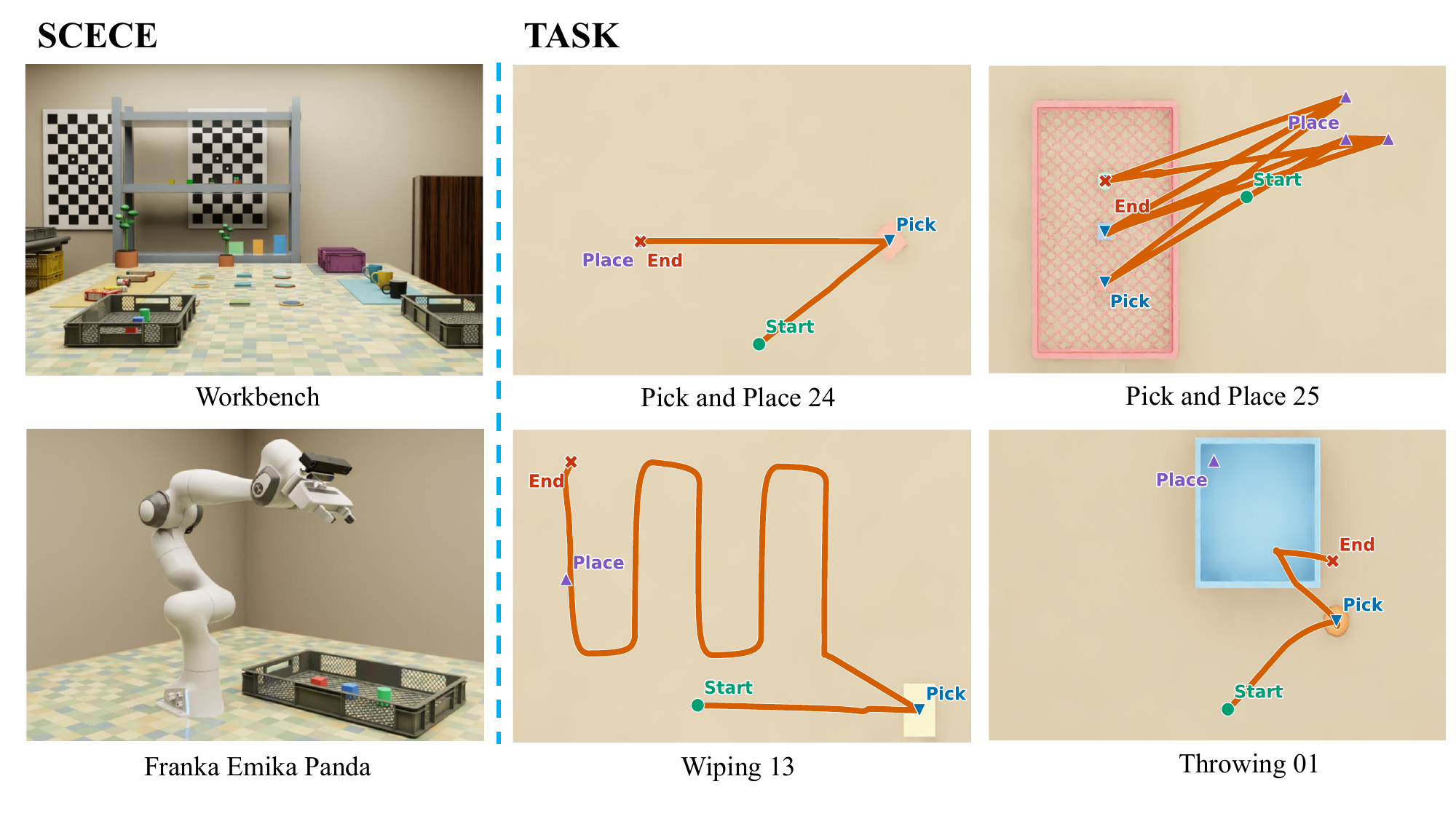}
\caption{\mildsim overview. \textbf{Scene:} Isaac Sim workbench and Franka manipulation setup. \textbf{Task:} representative pick-and-place, wiping, and throwing trajectories with executed end-effector paths.}
\label{fig:mildsim_overview}
\end{figure}

\mildsim uses Isaac Sim for pick-and-place, wiping, throwing, stacking, and insertion tasks with observable robot, object, and trajectory states (Fig.~\ref{fig:mildsim_overview}). Quantitative localization results in this paper come from the real-world subset, whereas \mildsim provides controlled replay episodes for the downstream diagnostic in Sec.~\ref{sec:exp_replay}.

\section{AprilVINS}
\label{sec:aprilvins}

\subsection{Overview}
\begin{figure*}[t]
\centering
\includegraphics[width=0.97\textwidth]{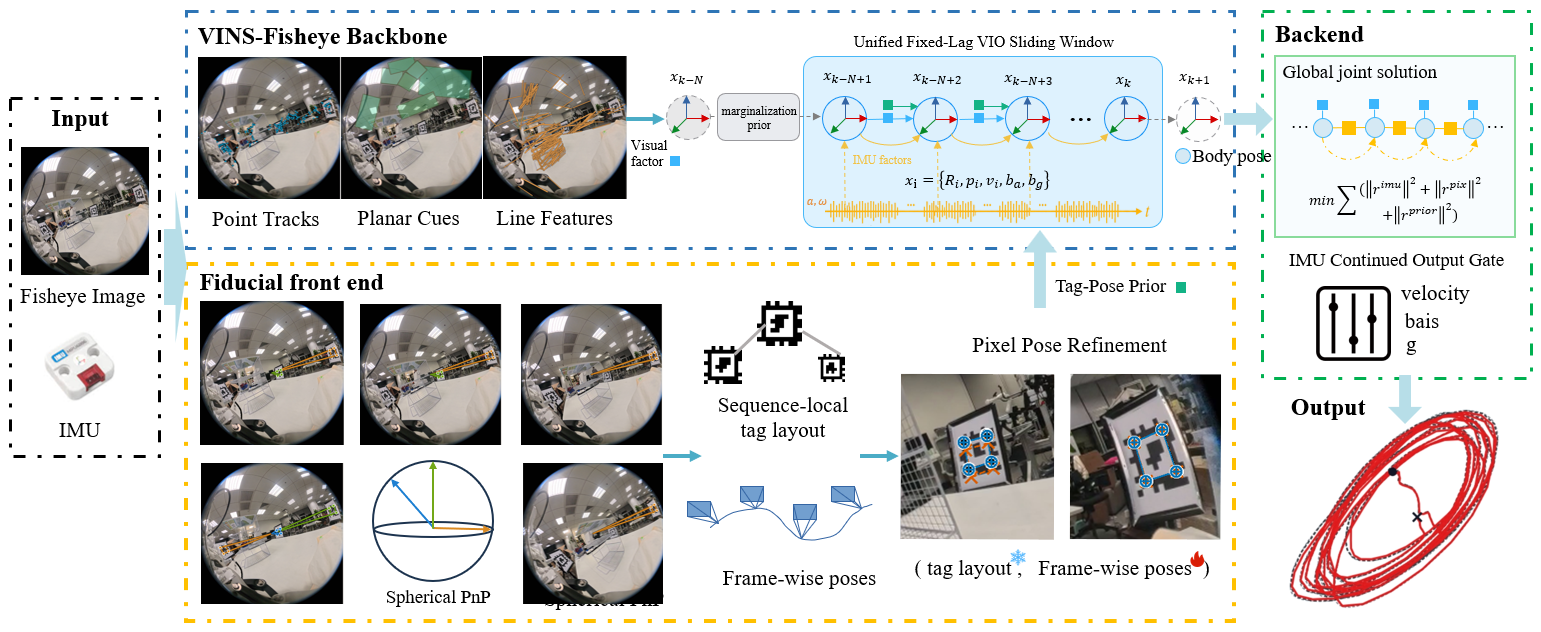}
\caption{\aprilvins pipeline. Fisheye images and IMU feed a VINS-Fisheye backbone with point, plane, and line features in a fixed-lag sliding window. A parallel fiducial front end detects AprilTags, estimates sequence-local tag layout via spherical PnP, and supplies admitted tag-pose priors to the window. Guarded continuation then filters the jointly optimized body state before TCP conversion.}
\label{fig:aprilvins_pipeline}
\end{figure*}

\aprilvins targets marker-augmented teaching workspaces without a pre-surveyed fiducial map by combining a fisheye visual--inertial backbone with a fiducial front end that estimates sequence-local tag geometry from known tag IDs and sizes. Admitted tag-pose priors enter a fixed-lag sliding window, and a separate guarded-continuation rule filters the joint state exported as the reported trajectory. A VIO run without tag factors is retained for ablation. Fig.~\ref{fig:aprilvins_pipeline} summarizes the data flow from fisheye/IMU inputs to the guarded TCP trajectory.

The design separates two decisions that are often conflated. Prior admission determines whether a tag observation may influence the VIO state, whereas guarded continuation determines whether the resulting joint state is stable enough to export. Thus, one accepted detection does not directly define the reported trajectory, and estimation accuracy can be analyzed separately from temporal continuity.

\subsection{State Representation}
We use a compact sliding-window representation aligned with the visual--inertial backbone. Let $\mathbf{T}_{AB}\in SE(3)$ map a point from frame $B$ to frame $A$. At keyframe $k$, the body state in the VIO frame $\mathcal{V}$ is
\begin{equation}
\mathbf{x}_k =
\{\mathbf{R}_{VB,k},\mathbf{p}_{VB,k},\mathbf{v}_{k},
\mathbf{b}_{a,k},\mathbf{b}_{g,k}\},
\label{eq:aprilvins_state}
\end{equation}
where $\mathbf{R}_{VB,k}$ and $\mathbf{p}_{VB,k}$ are the body orientation and position, $\mathbf{v}_{k}$ is velocity, and $\mathbf{b}_{a,k},\mathbf{b}_{g,k}$ are IMU biases. The active variables are
\begin{equation}
\mathcal{X}=\{\mathbf{x}_{k-N+1},\ldots,\mathbf{x}_{k},\mathbf{T}_{MV}\},
\label{eq:aprilvins_window}
\end{equation}
where $\mathbf{T}_{MV}$ aligns the VIO frame with the sequence-local tag-map frame $\mathcal{M}$.

\subsection{Fisheye Tag Observation Model}
AprilVINS uses camera-frame unit-sphere bearings as the AprilTag observation model for fisheye images. For tag $j$, corner $\ell$, and camera $c$, the measured fisheye pixel $\mathbf{u}_{j\ell k}^{c}$ is back-projected through the calibrated inverse projection and normalized as
\begin{equation}
\hat{\mathbf{r}}_{j\ell k}^{c}
=
\frac{\Pi_c^{-1}(\mathbf{u}_{j\ell k}^{c})}
{\|\Pi_c^{-1}(\mathbf{u}_{j\ell k}^{c})\|},
\label{eq:aprilvins_liftsphere}
\end{equation}
where $\Pi_c^{-1}(\cdot)$ denotes the calibrated fisheye inverse projection. If $\mathbf{P}_{j\ell}^{T_j}$ is the known corner coordinate from the tag size and $\mathbf{T}_{MT_j}$ is the estimated pose of tag $j$ in the sequence-local tag map, the predicted camera ray is
\begin{equation}
\bar{\mathbf{r}}_{j\ell k}^{c}
=
\pi_s\!\left(
    \mathbf{T}_{BC_c}^{-1}\mathbf{T}_{MB,k}^{-1}
\mathbf{T}_{MT_j}\mathbf{P}_{j\ell}^{T_j}
\right),
\label{eq:aprilvins_tag_ray}
\end{equation}
with $\pi_s(\mathbf{p})=\mathbf{p}/\|\mathbf{p}\|$ and known camera-to-body extrinsic $\mathbf{T}_{BC_c}$. The corner residual is evaluated on the tangent plane of the observed ray:
\begin{equation}
\mathbf{r}^{\mathrm{ray},c}_{j\ell k}
=
\mathbf{B}(\hat{\mathbf{r}}_{j\ell k}^{c})
\left(\bar{\mathbf{r}}_{j\ell k}^{c}-\hat{\mathbf{r}}_{j\ell k}^{c}\right),
\label{eq:aprilvins_ray_residual}
\end{equation}
where $\mathbf{B}(\cdot)$ is an orthonormal tangent basis. Ray residuals avoid a pinhole approximation near the image periphery and keep wide-field marker measurements consistent with the panoramic camera model.

\subsection{Sequence-Local Tag Layout and Pose Prior}
\aprilvins does not assume a pre-surveyed fiducial-map prior. Given known tag IDs and physical tag sizes, its fiducial front end estimates a sequence-local tag layout and frame-wise body poses from observations available up to the current frame:
\begin{equation}
\{\hat{\mathbf{T}}_{MT_j},\hat{\mathbf{T}}_{MB,k}\}
=
\arg\min
\sum_{j,\ell,k,c}
\rho\!\left(
\left\|\mathbf{\Omega}_{j\ell k}^{c\,1/2}
\mathbf{r}^{\mathrm{ray},c}_{j\ell k}\right\|^2
\right),
\label{eq:aprilvins_tagmap}
\end{equation}
where $\rho(\cdot)$ is a robust loss and $\mathbf{\Omega}_{j\ell k}^{c}$ is the corner-ray information matrix. The first accepted tag fixes the gauge of $\mathcal{M}$, and subsequent optimization uses only the connected observation graph available at frame $k$ without future frames. Users can therefore place tags without measuring their global poses.

The solved tag-based body pose $\hat{\mathbf{T}}_{MB,k}$ is then used as a gated prior on the VIO state and map--VIO alignment. The residual is
\begin{equation}
\mathbf{r}_{p,k}
=
\mathbf{\Sigma}_{p}^{-1/2}
\left(
\mathbf{p}_{MV}+\mathbf{R}_{MV}\mathbf{p}_{VB,k}
-\hat{\mathbf{p}}_{MB,k}
\right),
\label{eq:aprilvins_position_prior}
\end{equation}
\begin{equation}
\mathbf{r}_{R,k}
=
\mathbf{\Sigma}_{R}^{-1/2}
\operatorname{Log}
\left(
\hat{\mathbf{R}}_{MB,k}^{\top}
\mathbf{R}_{MV}\mathbf{R}_{VB,k}
\right),
\label{eq:aprilvins_rotation_prior}
\end{equation}
where $\mathbf{\Sigma}_{p}$ and $\mathbf{\Sigma}_{R}$ encode translational and rotational confidence. The tag-prior loss is
\begin{equation}
\mathcal{L}_{\mathrm{tag}}^k
=
\rho\!\left(
\left\|
\begin{bmatrix}
\mathbf{r}_{p,k}\\
\mathbf{r}_{R,k}
\end{bmatrix}
\right\|^2
\right).
\label{eq:aprilvins_tag_prior_loss}
\end{equation}

\subsection{Tag-Prior Admission and Guarded Output}
The tag-pose prior is compatible with the visual--inertial sliding-window objective:
\begin{equation}
\min_{\mathcal{X}}
\mathcal{L}_{\mathrm{prior}}
+
\sum_{k\in\mathcal{W}}
\left(
\mathcal{L}_{\mathrm{imu}}^k
+
\mathcal{L}_{\mathrm{vis}}^k
+
a_k^{\mathrm{tag}}\lambda_{\mathrm{tag}}\mathcal{L}_{\mathrm{tag}}^k
\right),
\label{eq:aprilvins_objective}
\end{equation}
where $a_k^{\mathrm{tag}}\in\{0,1\}$ is the prior-insertion gate that adds a tag-pose prior only when the synchronized tag pose passes geometric, timestamp, and SE(3)-consistency checks. This gate modifies the residuals in the existing fixed-lag sliding window rather than creating a second backend optimizer. After admitted tag factors are formed, AprilVINS uses a guarded continuation rule to decide which optimized state is safe to export. Each stage starts from the previously accepted state and uses a predefined inertial-weight schedule. Let $s_j$ denote the solver success flag at stage $j$. The accepted state is
\begin{equation}
\begin{aligned}
j^*&=\max\{j:\ s_1=\cdots=s_j=1\},\\
\hat{\mathcal{X}}&=\mathcal{X}^{(j^*)},\\
\hat{\mathbf{T}}_{MB,k}^{\mathrm{AV}}
&=\hat{\mathbf{T}}_{MV}\hat{\mathbf{T}}_{VB,k}.
\end{aligned}
  \label{eq:aprilvins_guard}
\end{equation}
If no stage succeeds, no sample is exported. Otherwise, the reported pose is composed from the last accepted tag-assisted visual--inertial solution \(\hat{\mathcal{X}}\).

An accepted output sample is therefore not a raw marker-only pose, but a state exported from the accepted joint solution. When tag factors are disabled, the same continuation rule still filters unstable VIO-only states before export. The two gating decisions play complementary roles: rejecting an inconsistent prior protects the sliding-window state from a spurious tag pose, whereas rejecting a failed continuation stage prevents an unstable optimized state from being exported as a valid teaching trajectory. Because these decisions can affect temporal density without necessarily changing aggregate RMSE, Sec.~\ref{sec:experiments} evaluates APE RMSE together with time completion, and Sec.~\ref{sec:exp_ablation} further analyzes inter-sample continuity.

\section{Experiments}
\label{sec:experiments}

Evaluation is organized around five questions:
\begin{itemize}
\item \textbf{Q1. Task-Level Benchmarking:} How do natural-feature, marker-based, and visual--inertial routes compare across manipulation tasks on \mild?
\item \textbf{Q2. Marker-Configuration Sensitivity:} Does increasing marker count reliably improve baseline localization?
\item \textbf{Q3. \aprilvins Design:} Where do the accuracy and continuity gains of \aprilvins come from?
\item \textbf{Q4. Practical Teaching Routes:} How does \aprilvinsfull compare with commercial teaching-device trajectories?
\item \textbf{Q5. Downstream Utility:} How do reported localization errors compare with task-specific replay tolerance?
\end{itemize}
Studies in Secs.~\ref{sec:exp_task_level}--\ref{sec:exp_practical} characterize localization under the \mild protocol; Sec.~\ref{sec:exp_replay} supplies task-specific tolerance references in \mildsim for interpreting the reported errors. Unless noted otherwise, the main quantitative comparisons use Insta360 X5 AprilTag4 recordings. Each result reports APE RMSE (m), temporal coverage ratio (time completion), and continuity against the executed TCP reference through the calibrated end-effector chain, using SE(3) alignment without scale fitting and a 0.05\,s association tolerance. Time completion is the estimated timestamp span clipped to the reference interval, divided by the reference duration. Continuity is summarized by the maximum inter-sample timestamp gap on the exported trajectory; runs with gaps larger than 0.5\,s are marked $^\dagger$. Low RMSE on a short span must therefore be read jointly with coverage and continuity. Crosses ($\times$) denote runs without an evaluable trajectory. Route comparisons are diagnostic case studies under their respective recording setups, not a single controlled ranking. For \aprilvinsfull, all reported numbers follow a unified protocol in which each sequence uses a profile selected by the same rule set over gate thresholds, tag-prior weights, and continuation schedules; profile manifests will be released with the benchmark. Table~\ref{tab:route_definitions} summarizes the evaluated routes.

RMSE measures agreement only over associated samples and can therefore appear favorable when an estimator returns a short, easy segment. Time completion and continuity respectively expose missing temporal extent and internal fragmentation. We consequently regard low error as useful only when coverage is high and the trajectory remains sufficiently continuous for replay.

\begin{table}[t]
\centering
\caption{Evaluated \aprilvins routes in Sec.~\ref{sec:experiments}.}
\label{tab:route_definitions}
\scriptsize
\setlength{\tabcolsep}{3pt}
\renewcommand{\arraystretch}{1.05}
\adjustbox{width=\columnwidth,center}{%
\begin{tabular}{lccc}
\toprule
\textbf{Route} & \textbf{Tag front end} & \textbf{Tag factors in VIO} & \textbf{Output gate} \\
\midrule
\aprilvinsnotag & on & off & on \\
\aprilvinsnopromo & on & on & off \\
\aprilvinsfull & on & on & on \\
\bottomrule
\end{tabular}%
}
\vskip 0.5mm
{\raggedright\scriptsize For \aprilvinsnotag, the output gate filters unstable VIO-only states because tag priors are disabled.\par}
\end{table}

\subsection{Task-Level Benchmarking}
\label{sec:exp_task_level}
Table~\ref{tab:task_route_matrix} compares 15 Insta360 X5 tasks. Most methods use AprilTag4 recordings, while UcoSLAM and OpenVINS(ArUco) use matched ArUco4 recordings. TagSLAM(single-anchor) uses one fixed tag without an externally surveyed multi-tag map. The table exposes route-specific failure modes under close-range fisheye manipulation rather than a single matched ranking.

\begin{table*}[t]
\centering
\caption{Task-level localization on MILD: APE RMSE (m) / time completion.}
\label{tab:task_route_matrix}
\scriptsize
\setlength{\tabcolsep}{2pt}
\renewcommand{\arraystretch}{1.08}
\resizebox{\textwidth}{!}{%
\begin{tabular}{lccccccccccccccc}
\toprule
\textbf{Route} & \textbf{Analemma} & \textbf{Circular} & \textbf{Zigzag} & \textbf{\makecell{Bookshelf\\01}} & \textbf{\makecell{Bookshelf\\02}} & \textbf{\makecell{Box\\01}} & \textbf{\makecell{Box\\02}} & \textbf{\makecell{Pick-and-\\place 01}} & \textbf{\makecell{Pick-and-\\place 02}} & \textbf{\makecell{Pick-and-\\place 03}} & \textbf{\makecell{Pick-and-\\place 04}} & \textbf{\makecell{Pick-and-\\place 05}} & \textbf{\makecell{Pick-and-\\place 06}} & \textbf{\makecell{Wiping\\01}} & \textbf{\makecell{Wiping\\02}} \\
\midrule
VINS-Fisheye & 0.0818/0.98 & 0.0385/0.98 & 0.0526/0.99 & 0.0598/0.99 & 0.0803/1.00 & 0.0503/0.99 & 0.0671/0.87 & \thirdbest{0.0423/0.99} & 0.0314/0.99 & 0.0284/0.99 & 0.0691/0.99 & 0.0433/0.92 & 0.0349/0.99 & 0.0264/1.00 & 0.0358/0.99 \\
DM-VIO & \thirdbest{0.0504/1.00}\textsuperscript{$\dagger$} & \thirdbest{0.0328/1.00}\textsuperscript{$\dagger$} & $\times$ & 0.2399/0.97 & 591.5738/0.99 & 0.1425/1.00\textsuperscript{$\dagger$} & $\times$ & 0.1479/0.98 & 0.1064/0.98 & 0.7871/1.00\textsuperscript{$\dagger$} & 0.3276/1.00\textsuperscript{$\dagger$} & $\times$ & 0.2939/1.00\textsuperscript{$\dagger$} & $\times$ & $\times$ \\
ORB-SLAM3 & $\times$ & $\times$ & $\times$ & $\times$ & $\times$ & $\times$ & $\times$ & $\times$ & $\times$ & $\times$ & $\times$ & $\times$ & $\times$ & $\times$ & $\times$ \\
OpenVINS & 3.0149/0.78 & $\times$ & 60.7054/0.88 & 2.9991/0.91 & 29.7722/0.92 & 1.0523/0.92 & 43.3075/0.96 & 2.6814/0.89 & 2.6765/0.88 & 3.4719/0.89 & 11.4182/0.94 & 18.3236/0.96 & 6.8308/0.94 & 17.7003/0.94 & 10.2853/0.95 \\
TagSLAM(single-anchor) & 0.0703/1.00 & 0.0404/1.00 & 0.3199/1.00 & \thirdbest{0.0347/1.00} & 0.0608/1.00 & 0.1104/1.00 & 0.0896/1.00 & 0.0553/1.00 & \thirdbest{0.0286/1.00} & 0.0404/1.00 & 0.2516/1.00 & 0.4332/1.00 & \thirdbest{0.0304/1.00} & \thirdbest{0.0260/1.00} & 0.0947/1.00 \\
UcoSLAM & 4.4863/0.92\textsuperscript{$\dagger$} & 0.4330/0.96\textsuperscript{$\dagger$} & 0.3402/0.43 & $\times$ & 0.6350/0.17 & 0.9210/0.98\textsuperscript{$\dagger$} & 0.3342/0.56 & $\times$ & 0.3035/0.98 & 0.3684/0.98 & 0.7185/0.99\textsuperscript{$\dagger$} & 0.7328/0.64 & $\times$ & 0.2856/0.99 & 0.3525/0.72\textsuperscript{$\dagger$} \\
OpenVINS(ArUco) & 2.7249/0.78 & $\times$ & 10.6777/0.88 & 2.8848/0.90 & 4.6028/0.92 & 2.1863/0.92 & 0.5260/0.18 & $\times$ & 2.8176/0.89 & 5.0842/0.89 & 5.0162/0.61 & 18.6128/0.95 & 6.5439/0.94 & 0.3838/0.33 & 16.7636/0.95 \\
\midrule
\aprilvinsnotag & 0.0760/0.98 & 0.0424/0.98 & \thirdbest{0.0520/0.99} & 0.0532/0.99 & \thirdbest{0.0600/0.99} & \thirdbest{0.0500/0.99} & \thirdbest{0.0605/1.00} & 0.0426/0.99 & 0.0301/0.99 & \thirdbest{0.0279/0.99} & \thirdbest{0.0597/0.99} & \thirdbest{0.0393/1.00} & 0.0360/0.99 & 0.0266/0.99 & \thirdbest{0.0253/0.99} \\
\aprilvinsnopromo & \secondbest{0.0081/1.00}\textsuperscript{$\dagger$} & \secondbest{0.0067/1.00}\textsuperscript{$\dagger$} & \secondbest{0.0071/1.00}\textsuperscript{$\dagger$} & \secondbest{0.0036/1.00}\textsuperscript{$\dagger$} & \secondbest{0.0087/1.00}\textsuperscript{$\dagger$} & \secondbest{0.0062/1.00}\textsuperscript{$\dagger$} & \best{0.0091/1.00}\textsuperscript{$\dagger$} & \secondbest{0.0034/1.00}\textsuperscript{$\dagger$} & \secondbest{0.0038/1.00} & \best{0.0017/1.00} & \secondbest{0.0055/1.00} & \secondbest{0.0050/1.00}\textsuperscript{$\dagger$} & \best{0.0027/1.00} & \secondbest{0.0023/1.00} & \best{0.0031/1.00}\textsuperscript{$\dagger$} \\
\aprilvinsfull & \best{0.0049/1.00} & \best{0.0053/1.00} & \best{0.0059/1.00} & \best{0.0031/1.00} & \best{0.0076/1.00} & \best{0.0051/1.00} & \secondbest{0.0098/1.00} & \best{0.0032/1.00} & \best{0.0023/1.00} & \best{0.0017/1.00} & \best{0.0044/1.00} & \best{0.0036/1.00} & \secondbest{0.0028/1.00} & \best{0.0021/1.00} & \secondbest{0.0033/1.00} \\
\bottomrule
\end{tabular}
}
\vskip 1mm
{\raggedright\scriptsize $^\dagger$ maximum inter-sample timestamp gap is larger than 0.5\,s.\par}
\end{table*}

\begin{figure}[t]
\centering
\includegraphics[width=\columnwidth]{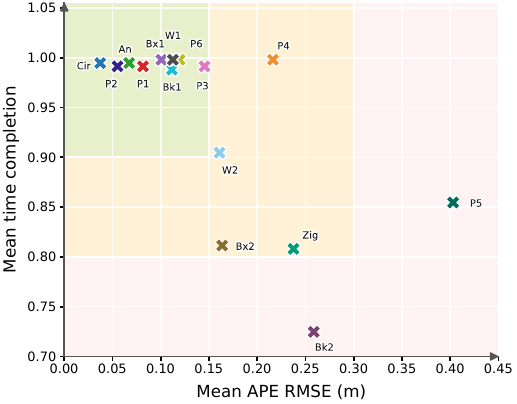}
\caption{Task-difficulty map from non-\aprilvins baselines. Points show the mean APE RMSE and mean time completion of the three lowest-RMSE evaluable routes per task.}
\label{fig:task_difficulty_map}
\end{figure}

Under the tested AprilTag4 recordings, marker-free fisheye VIO remains near 30--80~mm RMSE with high completion, whereas \aprilvinsfull reports 2--10~mm on the same sequences. Several baselines leave tasks without evaluable trajectories or with post-initialization divergence under close-range fisheye motion (Table~\ref{tab:task_route_matrix}). DM-VIO can also exhibit catastrophic RMSE spikes under viewpoint change, and routes marked with $^\dagger$ remain fragmented despite high completion. ORB-SLAM3 yields no evaluable trajectory on any listed task, which suggests that pinhole-oriented SLAM backbones transfer poorly to close-range fisheye manipulation without additional adaptation. Fig.~\ref{fig:task_difficulty_map} summarizes this spread: harder tasks occupy the upper-left (high RMSE, low completion), while pick-and-place and wiping cluster toward lower error when at least one route remains evaluable. Bookshelf and Box entries are where DM-VIO, OpenVINS, and UcoSLAM most often diverge, whereas TagSLAM(single-anchor) can track continuously yet still report 25--43~cm RMSE on Pick-and-place~04--05 and Wiping~02. These magnitudes are contextualized in Sec.~\ref{sec:exp_replay}.

\subsection{Marker-Configuration Sensitivity}
\label{sec:exp_marker}
Table~\ref{tab:marker_configuration} compares two-marker and four-marker results on six Insta360 X5 tasks (AprilTag2/4 or ArUco2/4) as a configuration-sensitivity study.

\begin{table}[t]
\centering
\caption{Marker-configuration sensitivity on Insta360 X5: APE RMSE (m) / time completion.}
\label{tab:marker_configuration}
\scriptsize
\setlength{\tabcolsep}{2pt}
\renewcommand{\arraystretch}{1.06}
\resizebox{\columnwidth}{!}{%
\begin{tabular}{llcccccc}
\toprule
\textbf{Method} & \textbf{Markers} & \textbf{Analemma} & \textbf{Circular} & \textbf{Zigzag} & \textbf{\makecell{Bookshelf\\01}} & \textbf{\makecell{Bookshelf\\02}} & \textbf{\makecell{Wiping\\02}} \\
\midrule
UcoSLAM & ArUco2 & $\times$ & 0.027/0.02 & 0.251/0.37 & 0.237/0.98 & 0.574/0.19 & $\times$ \\
 & ArUco4 & $\times$ & 0.433/0.96 & 0.340/0.43 & $\times$ & 0.635/0.17 & 0.352/0.72 \\
\midrule
OpenVINS(ArUco) & ArUco2 & $\times$ & $\times$ & $\times$ & 3.050/0.90 & 4.826/0.93 & $\times$ \\
 & ArUco4 & 2.725/0.78 & $\times$ & $\times$ & 2.885/0.90 & 4.603/0.92 & $\times$ \\
\midrule
TagSLAM(actual-count) & AprilTag2 & 0.063/1.00 & $\times$ & 0.072/1.00 & 0.035/1.00 & $\times$ & 0.026/1.00 \\
 & AprilTag4 & $\times$ & $\times$ & $\times$ & $\times$ & $\times$ & $\times$ \\
\midrule
\aprilvinsfull & AprilTag2 & 0.007/0.93 & 0.009/0.93 & 0.019/0.98 & 0.005/0.97 & 0.012/0.19 & 0.011/0.98 \\
 & AprilTag4 & 0.0049/1.00 & 0.0053/1.00 & 0.0059/1.00 & 0.0031/1.00 & 0.0076/1.00 & 0.0033/1.00 \\
\bottomrule
\end{tabular}
}
\end{table}

\begin{table*}[t]
\centering
\caption{Teaching-device trajectories: APE RMSE (m) / time completion.}
\label{tab:aprilvins_practical_matrix}
\scriptsize
\setlength{\tabcolsep}{2pt}
\renewcommand{\arraystretch}{1.08}
\resizebox{\textwidth}{!}{%
\begin{tabular}{lccccccccccc}
\toprule
\textbf{Route} & \textbf{Analemma} & \textbf{Circular} & \textbf{Zigzag} & \textbf{\makecell{Bookshelf\\01}} & \textbf{\makecell{Bookshelf\\02}} & \textbf{\makecell{Box\\01}} & \textbf{\makecell{Pick-and-\\place 04}} & \textbf{\makecell{Pick-and-\\place 05}} & \textbf{\makecell{Pick-and-\\place 06}} & \textbf{\makecell{Wiping\\01}} & \textbf{\makecell{Wiping\\02}} \\
\midrule
\textit{Body/waist motion} & No & No & Yes & Yes & No & Yes & Yes & Yes & Yes & Yes & Yes \\
\midrule
Insight9 SDK VIO & 0.097/1.00 & 0.015/1.00 & 0.014/1.00 & 0.034/1.00 & 0.063/1.00 & 0.565/1.00 & 0.019/1.00 & 0.025/1.00 & 0.023/1.00 & 0.201/1.00 & 0.029/1.00 \\
PICO Controller & 0.043/1.00 & 0.019/1.00 & 0.077/1.00 & 0.140/1.00 & 0.071/1.00 & 0.052/1.00 & 0.043/1.00 & 0.029/1.00 & 0.045/1.00 & 0.025/1.00 & 0.026/1.00 \\
\aprilvinsfull & \best{0.0049/1.00} & \best{0.0053/1.00} & \best{0.0059/1.00} & \best{0.0031/1.00} & \best{0.0076/1.00} & \best{0.0051/1.00} & \best{0.0044/1.00} & \best{0.0036/1.00} & \best{0.0028/1.00} & \best{0.0021/1.00} & \best{0.0033/1.00} \\
\bottomrule
\end{tabular}
}
\end{table*}

Increasing marker count alone does not reliably improve localization: TagSLAM(actual-count) yields AprilTag2 trajectories on four of six tasks but none on AprilTag4, and UcoSLAM can show lower RMSE with ArUco2 yet near-zero completion on Circular. The ArUco2 Circular entry (0.027/0.02) is especially misleading because millimeter RMSE on a 2\% completed span would still fail replay. \aprilvinsfull nonetheless improves on every task from AprilTag2 to AprilTag4 and restores Bookshelf~02 completion from 0.19 to 1.00. Table~\ref{tab:marker_configuration} therefore highlights how marker layout interacts with prior generation and output continuity, not marker count alone. Bookshelf~02 under AprilTag2 is illustrative: RMSE stays near 12~mm but completion drops to 0.19, so a teaching trajectory can remain unusable despite apparently low error. Sec.~\ref{sec:exp_ablation} next isolates which \aprilvins components recover both accuracy and continuity.

\subsection{\aprilvins Design Ablations}
\label{sec:exp_ablation}
We compare three configurations from Table~\ref{tab:route_definitions}: VIO without tag factors (\aprilvinsnotag), tag-assisted estimation without guarded output (\aprilvinsnopromo), and the full guarded output (\aprilvinsfull). On the same four-tag tasks, \aprilvinsnotag stays at centimeter scale, whereas \aprilvinsnopromo and \aprilvinsfull reach millimeter-level aligned RMSE, so admitted tag factors drive most of the accuracy gain. Comparing \aprilvinsnopromo and \aprilvinsfull in Table~\ref{tab:task_route_matrix} shows that the two routes often report similar RMSE, so accuracy and exportability should be read separately under manipulation-interface localization.

\begin{figure}[t]
\centering
\includegraphics[width=0.48\columnwidth]{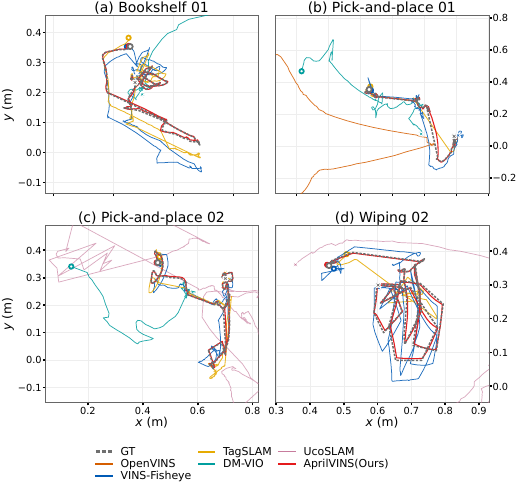}
\hfill
\includegraphics[width=0.48\columnwidth]{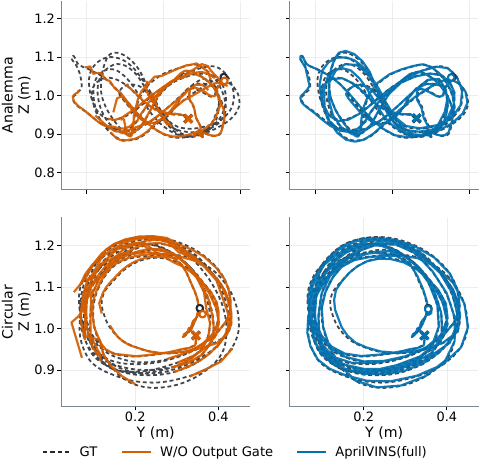}
\caption{AprilVINS ablation visualizations. \textbf{Left:} local trajectory comparisons on four \mild tasks; dashed curves are executed TCP references and solid curves are estimates. \textbf{Right:} output-gate continuity check in the YZ plane; curves split at timestamp gaps larger than 0.15\,s.}
\label{fig:aprilvins_ablation}
\end{figure}

On the same tasks, \aprilvinsnopromo retains $^\dagger$ fragmentation markers whereas \aprilvinsfull does not, which matches Fig.~\ref{fig:aprilvins_ablation}: close TCP tracking with the largest local departures near fast approach, and guarded output removing fragment breaks larger than 0.15~s even when span completion remains high. The marker-configuration and ablation results show that marker-augmented workspaces require reliable tag priors and stable export. Sec.~\ref{sec:exp_practical} next compares \aprilvinsfull with commercial teaching devices.

\subsection{Practical Teaching Routes}
\label{sec:exp_practical}
\begin{figure}[t]
\centering
\includegraphics[width=\columnwidth]{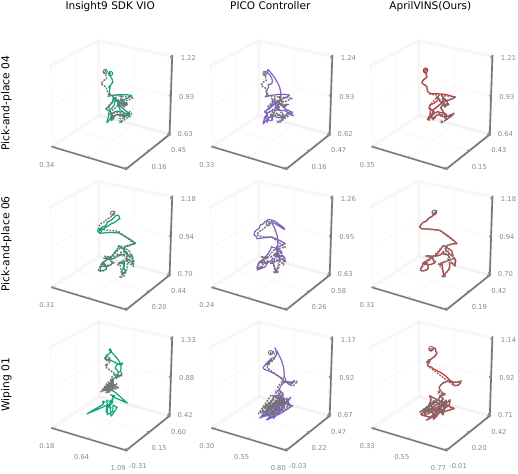}
\caption{3D practical-route trajectory comparisons for three Table~\ref{tab:aprilvins_practical_matrix} tasks. Dashed curves are route-specific executed references; solid curves are estimates.}
\label{fig:tablev_practical_route_9grid}
\end{figure}

Table~\ref{tab:aprilvins_practical_matrix} compares Insight9 SDK VIO, PICO Controller, and \aprilvinsfull on overlapping tasks recorded with different teaching hardware. Box~02 and Pick-and-place~01--03 are omitted because the released recordings do not provide a matched overlap set for all three routes. On the overlapping tasks listed, \aprilvinsfull reports the lowest aligned RMSE among the three routes with full time completion, but the comparison is not hardware-matched and should not be read as a controlled cross-device ranking. Insight9 SDK VIO remains strong on compact trajectories such as Circular and Pick-and-place~04--06, whereas PICO Controller shows larger offsets on tasks with body or waist motion. The body/waist-motion row aligns with this pattern: tasks marked ``Yes'' tend to show larger PICO offsets even when completion stays at 1.00. Insight9 SDK VIO on Box~01 (0.565~m) and Wiping~01 (0.201~m) shows that full completion can still coexist with locally replay-critical error. Fig.~\ref{fig:tablev_practical_route_9grid} illustrates tighter 3D tracking for \aprilvinsfull on three representative tasks. Sec.~\ref{sec:exp_replay} next relates these localization errors to task-level replay tolerance in \mildsim.

\begin{figure}[t]
    \centering
    \includegraphics[width=0.95\columnwidth]{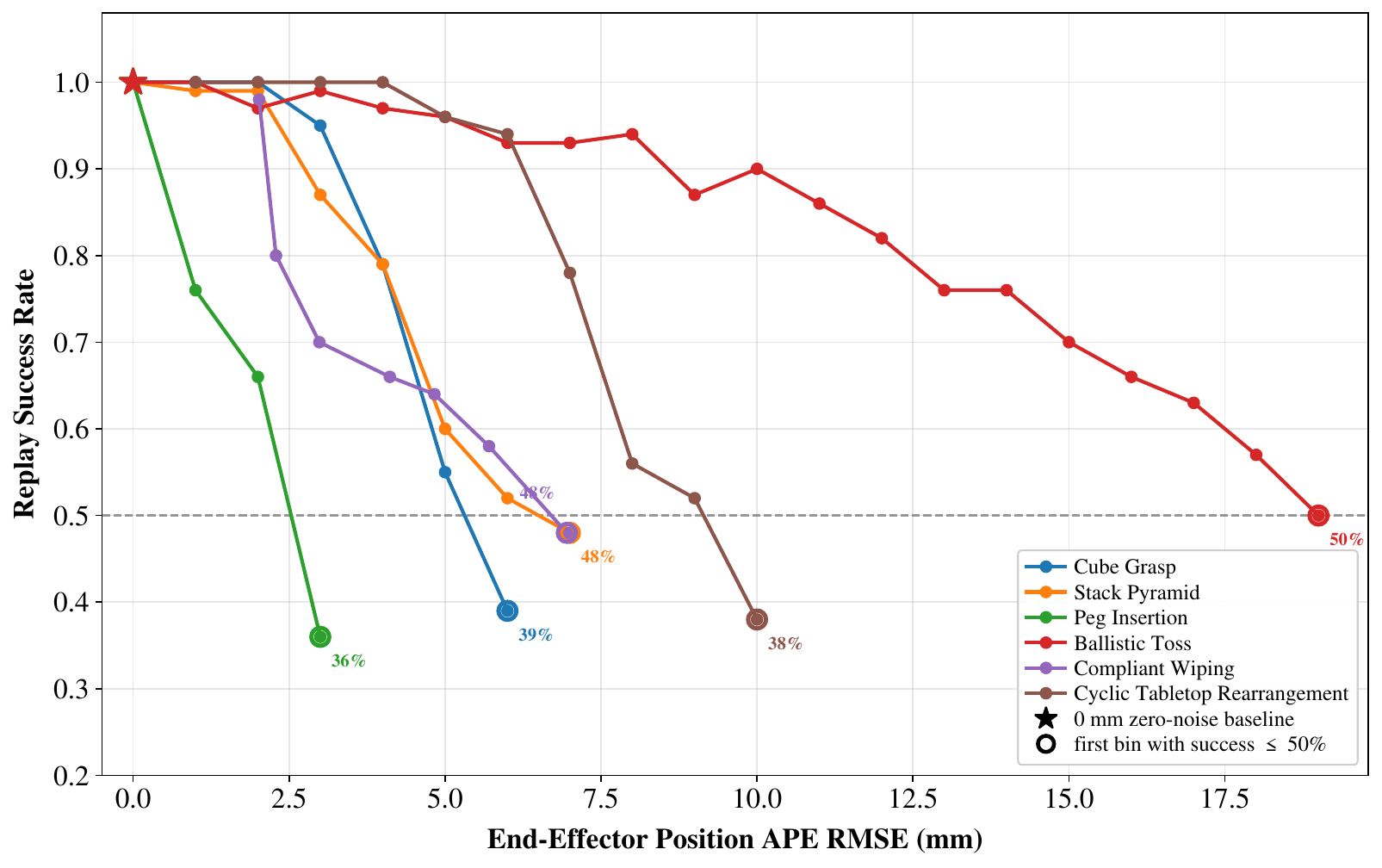}
    \caption{Replay success versus translational APE--RMSE on six \mildsim tasks. Dashed lines mark the zero-noise baseline and the first bin at or below 50\% success; error bars show Wilson 95\% confidence intervals.}
    \label{fig:ape-replay-success}
    \end{figure}

\subsection{Downstream Replay Study}
\label{sec:exp_replay}

Sec.~\ref{sec:introduction} motivates relating estimator error to task-level consequences. Rather than replaying the benchmarked trajectories, this study injects controlled translational perturbations into a fixed successful source episode to obtain task-specific tolerance curves. For each task, a successful source episode is fixed and replayed after adding zero-mean cubic B-spline translation noise to the TCP position targets, while orientations and gripper commands stay unchanged. Realized translational APE--RMSE is computed without spatial alignment:
\begin{equation}
    \mathrm{APE}_{\mathrm{RMSE}}^{\mathrm{pos}}
    = 10^{3}\sqrt{\frac{1}{T}\sum_{t=1}^{T}
        \left\|\mathbf{p}_{\mathrm{replay},t}
        - \mathbf{p}_{\mathrm{source},t}\right\|_{2}^{2}}
    \quad [\mathrm{mm}],
    \label{eq:replay_ape_rmse}
\end{equation}
where $\mathbf{p}_{\mathrm{source},t}$ and $\mathbf{p}_{\mathrm{replay},t}$ are expressed in meters. Each bin contains 100 valid trials whose measured APE--RMSE falls within $\left[c-0.5,c+0.5\right)$~mm, and task failures count as unsuccessful replays. Because orientations and gripper commands are held fixed, the study isolates translational execution error from other replay factors.

Fig.~\ref{fig:ape-replay-success} maps realized translational error to replay success for six \mildsim tasks. All curves start at 100\% success at zero noise and then fall beyond a task-specific knee. The dashed 50\% lines span 3~mm for Peg Insertion Side (36\% at 3~mm) to 19~mm for Ballistic Toss, with Cube Grasp/Lift, Stack Pyramid, and Compliant Wiping near 6--7~mm and Cyclic Tabletop Rearrangement near 10~mm (38\% at 10~mm). Peg Insertion Side and Stack Pyramid decline earliest under millimeter-scale lateral error, whereas Ballistic Toss tolerates larger offsets and Compliant Wiping plus Cyclic Tabletop Rearrangement fall near 7--10~mm knees. Cube Grasp/Lift sits between these regimes because grasp clearance is tighter than free transport but less alignment-critical than insertion. Relative to the localization results above, marker-free fisheye VIO errors of 30--80~mm fall in the steep region for several contact-sensitive tasks, whereas reported \aprilvinsfull errors of 2--10~mm lie closer to the corresponding tolerance knees. The replay study therefore supplies tolerance references for interpreting the reported metrics, rather than validating estimator outputs directly.

\section{Conclusion}
We presented \mild for evaluating manipulation-interface localization and \aprilvins for exploiting sequence-local AprilTag geometry in marker-augmented teaching workspaces without a pre-surveyed fiducial map. On Insta360 AprilTag4 recordings, \aprilvinsfull under a unified protocol with sequence-specific profiles reaches millimeter-level SE(3)-aligned TCP-relative APE RMSE with high time completion and lower reported error than the tested routes under their respective protocols, whereas VIO without tag factors remains at centimeter scale. Reporting accuracy, coverage, and continuity jointly shows that low RMSE alone is insufficient for replayable teaching trajectories and that export continuity must be evaluated separately from aligned error. The \mildsim replay study supplies task-specific tolerance references without replaying benchmarked trajectories; \mild and \aprilvins thus form a diagnostic framework for UMI-style demonstration collection.

\bibliographystyle{IEEEtrans}
\bibliography{root}

\end{document}